\documentclass[fleqn,10pt]{wlscirep}
\usepackage[utf8]{inputenc}
\usepackage[T1]{fontenc}
\usepackage{xurl}

\usepackage{dblfloatfix}

\title{Sparse bottleneck neural networks for exploratory non-linear visualization of Patch-seq data}

\author[1,2]{Yves Bernaerts}
\author[1,2,3,*]{Philipp Berens}
\author[1,*]{Dmitry Kobak}
\affil[1]{Institute for Ophthalmic Research, University of T\"ubingen, Germany}
\affil[2]{International Max Planck Research School for Intelligent Systems, T\"ubingen, Germany}
\affil[3]{Tübingen AI Center, University of T\"ubingen, T\"ubingen, Germany}
\affil[*]{philipp.berens@uni-tuebingen.de, dmitry.kobak@uni-tuebingen.de}

\twocolumn
\begin{abstract}
Patch-seq, a recently developed experimental technique, allows neuroscientists to obtain transcriptomic and electrophysiological information from the same neurons. Efficiently analyzing and visualizing such paired multivariate data in order to extract biologically meaningful interpretations has, however, remained a challenge. Here, we use sparse deep neural networks with and without a two-dimensional bottleneck to predict electrophysiological features from the transcriptomic ones using a group lasso penalty, yielding concise and biologically interpretable two-dimensional visualizations. In two large example data sets, this visualization reveals known neural classes and their marker genes without biological prior knowledge. We also demonstrate that our method is applicable to other kinds of multimodal data, such as paired transcriptomic and proteomic measurements provided by CITE-seq.
\end{abstract}

\begin{document}

\setcounter{secnumdepth}{0}
\renewcommand{\thefootnote}{\roman{footnote}}
\flushbottom
\maketitle
\thispagestyle{empty}

\section*{Introduction}

\begin{figure*}[t]
  \centering    
  \includegraphics[width=\linewidth]{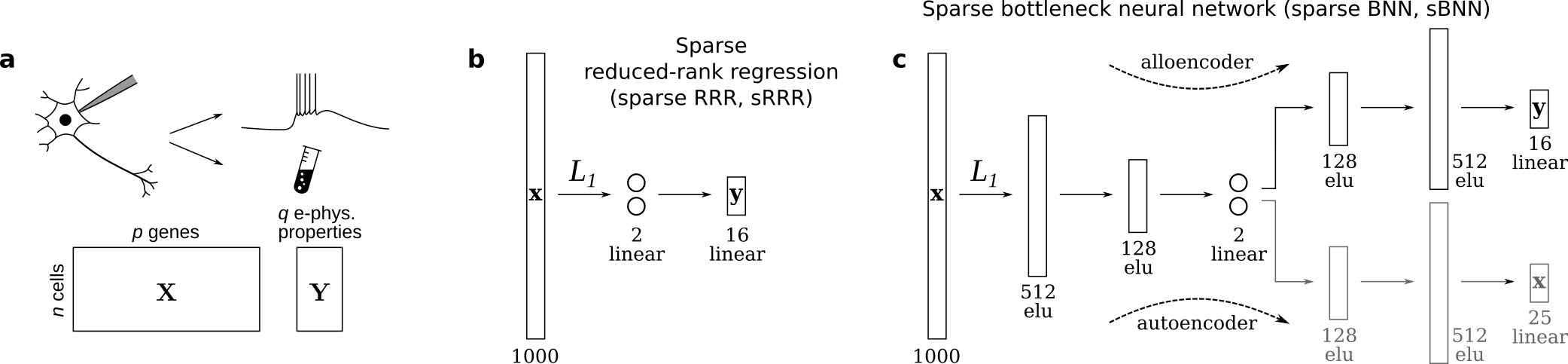}
  \caption{\textbf{(a)} Patch-seq allows for electrophysiological recordings and extraction of RNA content of the same cells. \textbf{(b)}~Sparse reduced-rank regression. \textbf{(c)}~Sparse bottleneck neural network.}
  \label{fig:schema}
\end{figure*}

\noindent 
Patch-seq is a recently developed experimental technique combining electrophysiological recordings and single-cell RNA sequencing (scRNA-seq) in the same set of neurons \cite{cadwell2016electrophysiological, cadwell2017multimodal, fuzik2016integration, foldy2016single}. Analyzing the resulting paired, or `two-view', datasets (Figure ~\ref{fig:schema}a) remains a challenge \cite{TripathyKobakPatchSeq}, as the dimensionality of the transcriptomic data is very high whereas the sample size is low due to the low-throughput nature of the technique.

There is a large number of linear and nonlinear methods for low-dimensional visualization of non-paired `one-view' data, including principal component analysis (PCA), t-distributed stochastic neighbor embedding (t-SNE) \cite{maaten2008visualizing}, and neural networks with `encoder-decoder' architecture called autoencoders \cite{hinton2006reducing, wollstadt20203Dpointcloud, demiralp2019data2vis}. Autoencoder networks have been employed for the analysis of scRNA-seq data \cite{yosef2018deep, theis2019deep}, showing the great potential of nonlinear parametric models for single-cell data analysis. However, very few such methods specifically focus on low-dimensional visualizations and scientific exploration of paired data. \cite{parviainen2010deep}.

We have previously developed a linear method called sparse reduced-rank regression (sRRR) \cite{kobak2019sparse} that linearly predicts electrophysiological features from gene expression via a low-dimensional bottleneck and employs a strong lasso penalty to perform feature selection (Figure~\ref{fig:schema}b). Here we introduce sparse bottleneck neural networks (sBNNs) --- a nonlinear generalization of sRRR (Figure~\ref{fig:schema}c) --- as a nonlinear framework for exploratory analysis and interpretable visualizations of paired datasets.

Our implementation and reproducible analysis are available at \url{https://github.com/berenslab/sBNN}).

\section*{Results}

\subsection*{Model performance}

\begin{figure}[t]
  \centering    
  \includegraphics[width=\linewidth]{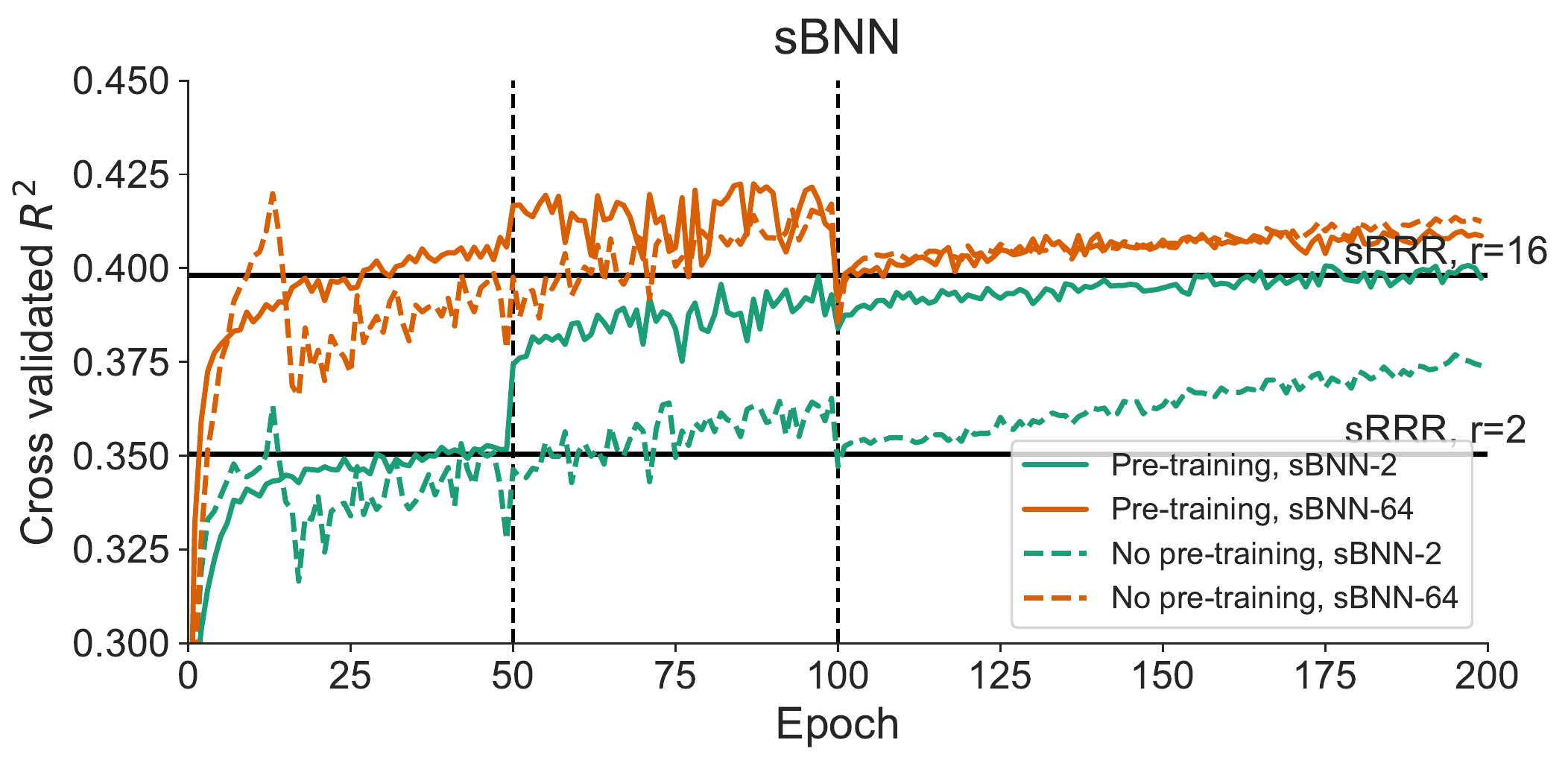}
  \caption{Cross-validated $R^2$ scores of the sparse bottleneck neural network with 2-and 64-dimensional bottleneck and with or without pretraining. The bottom two layers are frozen for 50 epochs after pre-training. All models were pruned to 25 input units at epoch 100 and trained for further 100 epochs. Horizontal lines show  performances of rank-2 and rank-16 sparse RRR with 25 genes.}
  \label{fig:training_curves}
\end{figure}

In our architecture (Figure~\ref{fig:schema}c), the sBNN compresses the high-dimensional transcriptomic data into a low-dimen\-sional bottleneck representation via several fully-connected layers, from which one head of the network reconstructs the transcriptomic data (`autoencoder'), whereas an additional head predicts the  electrophysiological properties of the neurons (`alloencoder'). We employed a group lasso penalty \cite{lin2006grouplasso} on the first encoding layer to enforce sparse gene selection \cite{wang2015grouplasso}, followed by pruning \cite{dally2015pruning, guttag2020pruning} of all input units except for a small predefined number (here, 25) and further fine-tuning. We also used a specialized training schedule consisting of classification-based pre-training \cite{lathuiliere2019comprehensive, transferlearning}, initial optimization with two frozen layers and subsequent optimization of all layers (see Methods). We found this training schedule to be advantageous for some datasets and model configurations (Figure~\ref{fig:training_curves}).

Our main focus is on data visualization and exploration, which is particularly convenient if the bottleneck representation is two-dimensional (as shown in Figure~\ref{fig:schema}c) and can be visualized directly. However, we also considered an architecture with a higher dimensional bottleneck in the latent space; for that we used a 64-dimensional layer (sBNN-64) instead of the 2-dimensional one (sBNN-2). In this case, we resort to t-SNE to make a two-dimensional visualization of the latent space.

We applied the sBNN to the two currently largest Patch-seq datasets: one dataset containing 1213 excitatory and inhibitory neurons from primary motor cortex (M1) of adult mice \cite{scala2020phenotypic} and another containing 3395 inhibitory neurons from primary visual cortex (V1) of adult mice \cite{gouwens2020toward}. The electrophysiological features used to characterize the physiology of each neuron contained features such as the action potential width and amplitude or the membrane time constant (see Methods for a complete list). Together, these two datasets broadly sampled the diversity of neural cell types in the neocortex \cite{scala2020phenotypic, gouwens2020toward}. 

To assess the predictive performance of the sBNN architectures we used 10-fold cross-validation (Table~\ref{table:sBNN_vs_sRRR_R2_scores}). For comparison, we also used our linear sRRR \cite{kobak2019sparse} that predicts electrophysiological features from gene expression via a linear bottleneck (Figure~\ref{fig:schema}b). The cross-validated $R^2$ of the sRRR model with a two-dimensional bottleneck (sRRR-2) with 25 genes was $0.35\pm 0.02$ (mean$\pm$SD across cross-validation folds) and $0.19\pm 0.01$ for the M1 and V1 datasets, respectively (Table~\ref{table:sBNN_vs_sRRR_R2_scores}). Without a 2D bottleneck (sRRR-full model), $R^2$ increased to $0.40$ and $0.25$ respectively. The sBNN-2's performance reached $0.40$ (Figure~\ref{fig:training_curves}, green line) and $0.26$ (Supplementary Figure~\ref{fig:training_curves_Allen}c) respectively. Importantly, sBNN-2 outperformed sRRR-2 models and performed as well as sRRR-full linear models. Increasing the dimensionality of the bottleneck (sBNN-64) led to further improvement of $R^2$ to 0.41 and 0.29 for the two datasets.

\begin{table}[b]
\centering
\begin{tabular}{llll}
\toprule
Model & M1 & V1 & CITE-Seq\\
\midrule
sRRR, rank-2 & $.35\pm .02$ & $.19\pm .01$ & $.23\pm .04$\\
sRRR, full-rank & $.40\pm .02$ & $.25\pm .01$ & $.35\pm .06$\\
sBNN-2 & $.40\pm .02$ & $.26\pm .01$ & $.38\pm .07$\\
sBNN-64 & $.41\pm .01$ & $.29\pm .01$ & $.41\pm .14$\\
\bottomrule
\end{tabular}
\caption{Cross-validated multivariate $R^2$ performance scores for the sRRR and sBNN frameworks regarding the M1, V1 and CITE-Seq data sets. Mean$\pm$SD across 10 cross-validation folds.}
\label{table:sBNN_vs_sRRR_R2_scores}
\end{table}

\begin{figure}[h!]
\centering
\includegraphics[width=\linewidth]{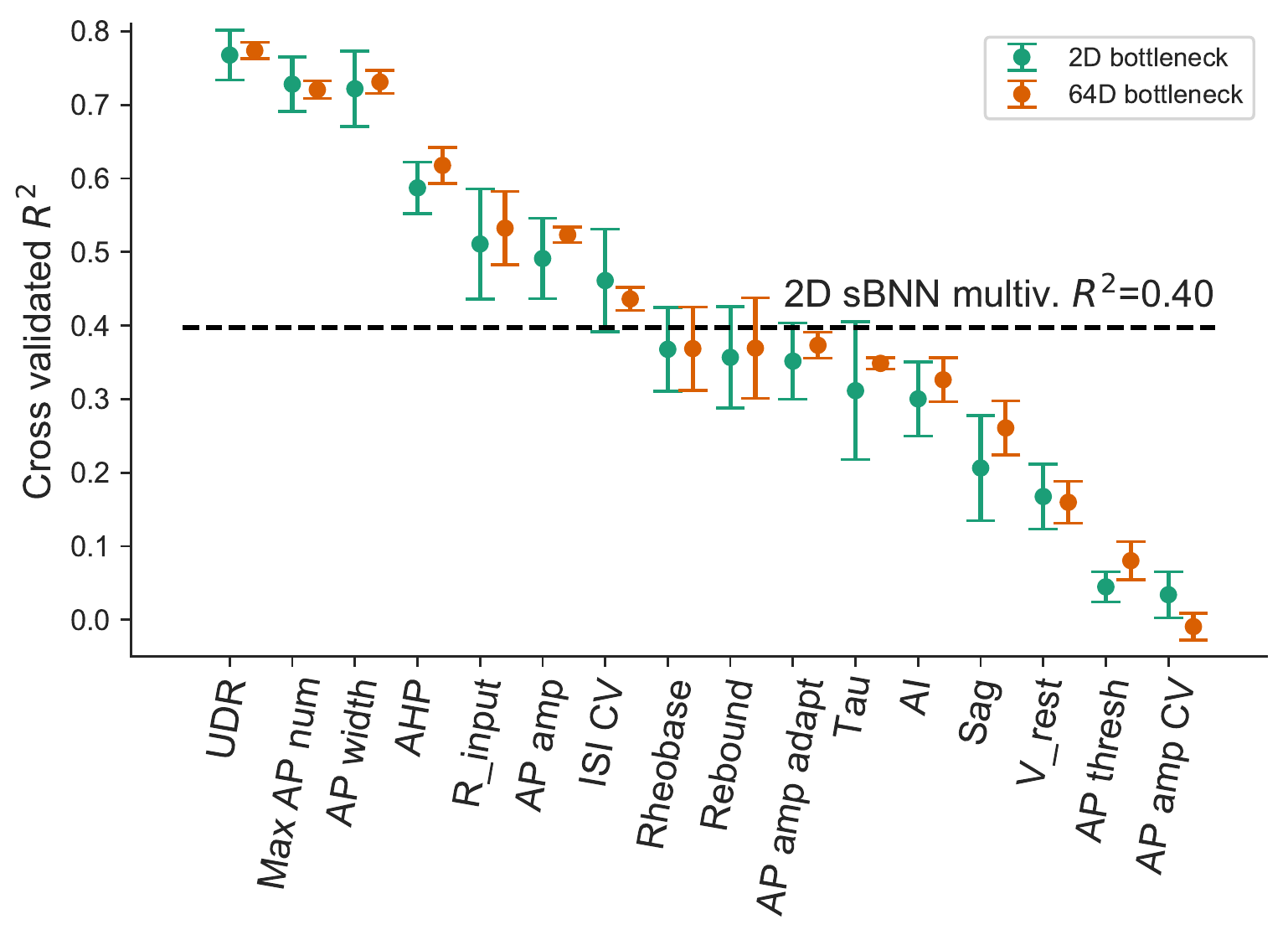}
\caption{Predictive performance for individual electrophysiological features in the M1 dataset using sBNN-2 and sBNN-64. Error bars show $\pm$SD across cross-validation folds. The horizontal line shows the multivariate cross-validated $R^2$ score for the model with a 2-dimensional bottleneck.}
\label{fig:ind_ephys_scores}
\end{figure}

Not all electrophysiological features could be predicted equally well. For the M1 dataset, the highest cross-validated $R^2$ values were obtained for the upstroke-downstroke ratio of the action potential ($R^2=0.77\pm 0.03$, sBNN-2), maximum action potential count ($R^2=0.73\pm 0.04$), and action potential width ($R^2=0.72\pm 0.05$), whereas some other features had $R^2$ below 0.1 (Figure~\ref{fig:ind_ephys_scores}). Using the sBNN-64 led to similar but slightly higher performances and  smaller variances across cross-validation folds (Figure~\ref{fig:ind_ephys_scores}).

\subsection*{Gene selection}
Having compared the predictive power of sBNN nonlinear models to sRRR linear ones, we next turned to analyzing the 25 genes selected by the group-lasso penalty to be used in the model. To this end, we retrained the sBNN models on the entire dataset.

\begin{figure}[ht]
\centering
\includegraphics[width=\linewidth]{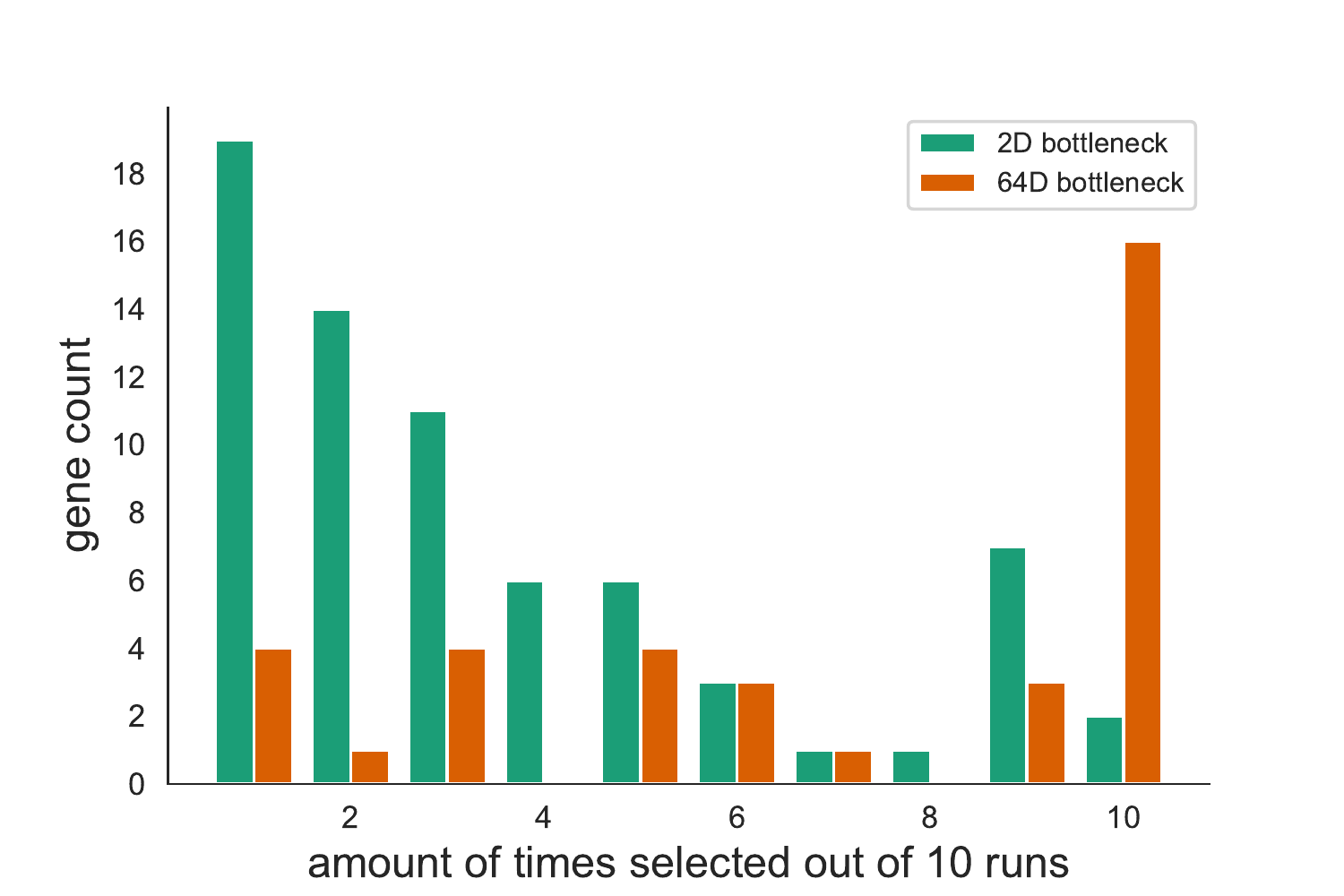}
\caption{Gene selection histogram showing how many genes were selected $k$ amount of times in trained sBNN-2 and sBNN-64 networks using 10 different random initializations.}
\label{fig:gene_histograms}
\end{figure}

Due to the sparseness penalty and the observed correlations between expression values for the different genes, the selected set of 25 genes was not the same in each model trained with different random initializations \cite{xu2011sparse}.  We found that this issue affected sBNN-2 more than sBNN-64: for the M1 dataset only two genes were selected on every run for sBNN-2, but 16 genes for sBNN-64 (Figure~\ref{fig:gene_histograms}). Similarly, 19 genes were selected on only one run for sBNN-2, but 4 genes on only one run for sBNN-64. 

For the M1 dataset and sBNN-64 model, among the selected genes we found \textit{Calb1, Sst, Vip, Pvalb, Gad1, Lamp5}, etc. (genes sorted in descending order by the $\ell_2$ norms of the first layer weights; see Table \ref{table:genes}), of which \textit{Sst, Vip, Pvalb}, and \textit{Lamp5} are well-known marker genes of specific neuronal populations \cite{tasic2018shared, trembley2016GABA}, where \textit{Gad1} encodes an enzyme responsible for catalyzing the production of gamma-aminobutyric acid (GABA), a crucial molecule in neuronal signaling and \textit{Calb1} encodes a protein thought to buffer entry of calcium upon stimulation of glutamate receptors \cite{rintoul2001calciumbuffering}. We can therefore conclude that sBNN-64 was very capable, without prior knowledge, to select biologically relevant and interpretable genes. For the V1 dataset and sBNN-64, the list started with \textit{Ptprt, Fxyd6, Npy, Cacna2d3, Brinp3, Vip}, etc. Notably here, \textit{Fxyd6} and \textit{Cacna2d3} encode important ion transport transmembrane proteins, directly affecting the electrophysiology of cells.

\begin{table}[h!]
\centering
\begin{tabular}{llp{5cm}}
\toprule
Data & Model & Genes \\
\midrule
M1 & sBNN-2 & \textit{Gad1, Pvalb, Erbb4, Slc6a1, Coro6, Tafa1, Ptk2b, Synpr, Cplx1, Gad2, Vip, Atp1a3, Fxyd6, Plch2, Npy1r, Calb1, Cbln2, Reln, Cplx3, Cacna1e, Parm1, Mybpc1, Col24a1, Elmo1, Rab3b} \\
\midrule
M1 & sBNN-64 & \textit{Calb1, Sst, Vip, Pvalb, Gad1, Lamp5, Fxyd6, Htr3a, Cplx1, Cbln2, Galnt14, Pdyn, Tac2, Ndn, Kcnc2, Elmo1, Reln, Erbb4, Ndst3, Tafa1, Enpp2, Gm49948, Gabra1, Plch2, Atp1b2} \\
\midrule
V1 & sBNN-2 & \textit{Vip, Synpr, Pdyn, Fxyd6, Trhde, Cacna2d3, Ptpru, Kcnc2, Pvalb, Hpse, Pcp4l1, Egfem1, Lamp5, Penk, Pde11a, Necab1, Chodl, Cpne7, Fstl4, Crtac1, Zfp536, Grm1, Shisa6, Kit, Grin3a}\\
\midrule
V1 & sBNN-64 & \textit{Ptprt, Fxyd6, Npy, Cacna2d3, Brinp3, Vip, Cacna1e, Flt3, Pvalb, Mybpc1, Synpr, Dpp10, Kcnab1, Penk, Trhde, Htr1f, Alk, Cdh7, Cntn4, Akr1c18, Sema5a, Syt6, Kctd8, Igf1, Unc5d}\\
\bottomrule
\end{tabular}
\caption{List of $25$ selected genes for M1 and V1 datasets, ranked in descending $\ell_2$ norm order of sBNN-2 and sBNN-64 first layer weights.}
\label{table:genes}
\end{table}

\subsection*{Latent space visualization}

\begin{figure*}[ht!]
\centering
\includegraphics[width=\linewidth]{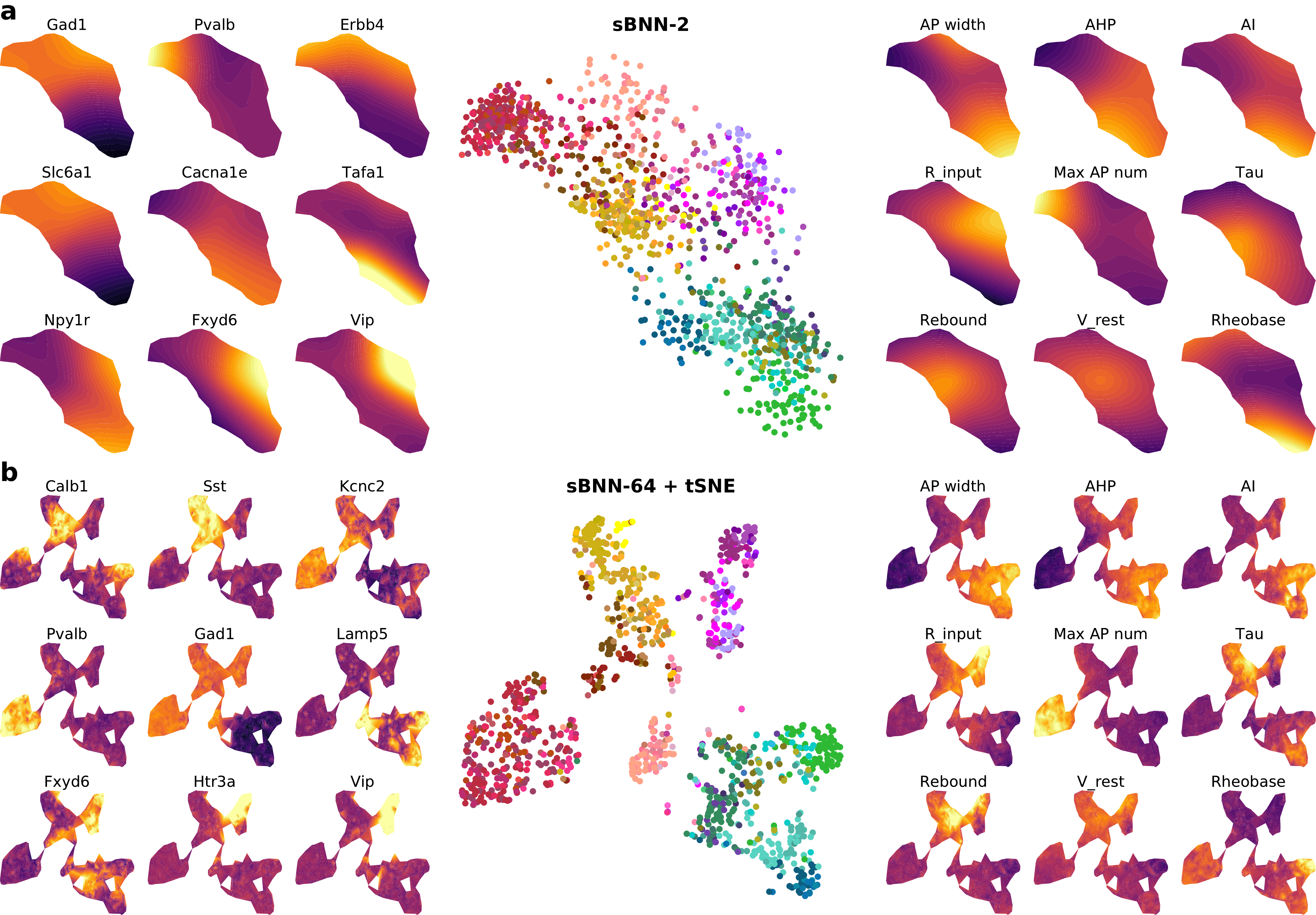}
\caption{The M1 dataset. \textbf{(a)} sBNN-2 model. Middle: embedded latent space of the sBNN model trained on the entire dataset (all $n=1213$ cells are shown). Colors correspond to transcriptomic types and are taken from the original publication \cite{scala2020phenotypic}. Red: \textit{Pvalb} interneurons; yellow: \textit{Sst} interneurons; purple: \textit{Vip} interneurons; salmon: \textit{Lamp5} interneurons; green/blue: excitatory neurons. Right: alloencoder model predictions for nine exemplary electrophysiological features. Left: autoencoder model predictions for nine exemplary genes out of the 25 selected genes. \textbf{(b)} sBNN-64 model. T-SNE was performed to embed the 64-dimensional latent space in a 2-dimensional visualization. Analagous to (a). See also Methods.}
\label{fig:scala_full_dataset}
\end{figure*}

\begin{figure}[ht!]
\centering
\includegraphics[width=\linewidth]{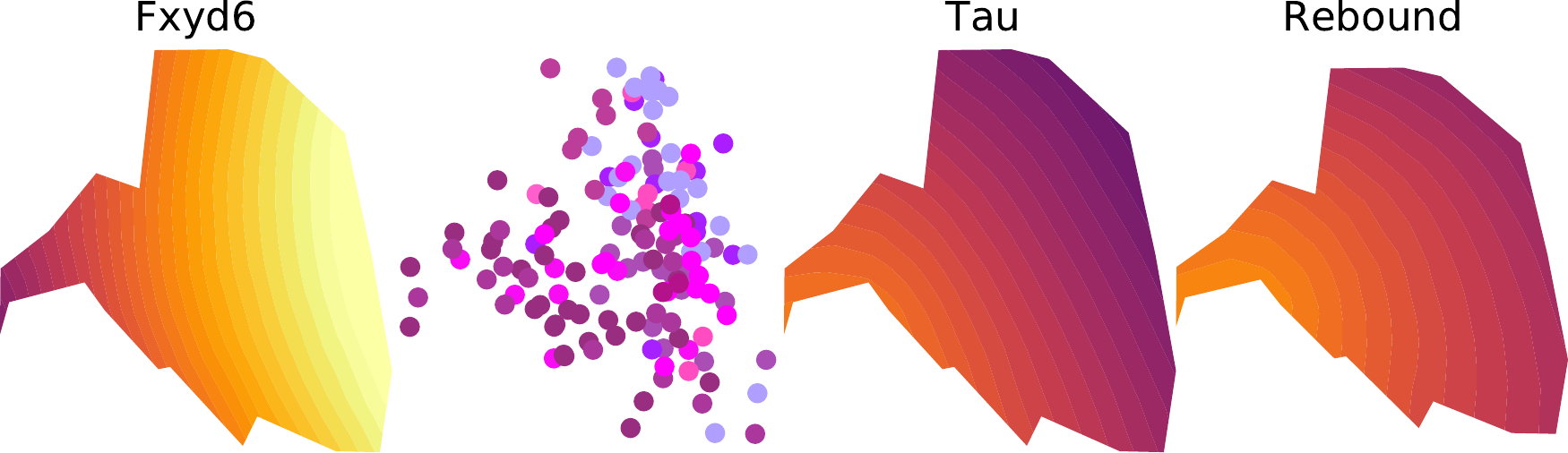}
\caption{Zoomed in version of Figure~\ref{fig:scala_full_dataset}b for \textit{Vip} neurons. The \textit{Fxyd6} ion channel gene, rebound and membrane time constant overlays are shown. See Methods.}
\label{fig:scala_full_dataset_zoomed_in}
\end{figure}

\begin{figure*}[ht]
\centering
\includegraphics[width=\linewidth]{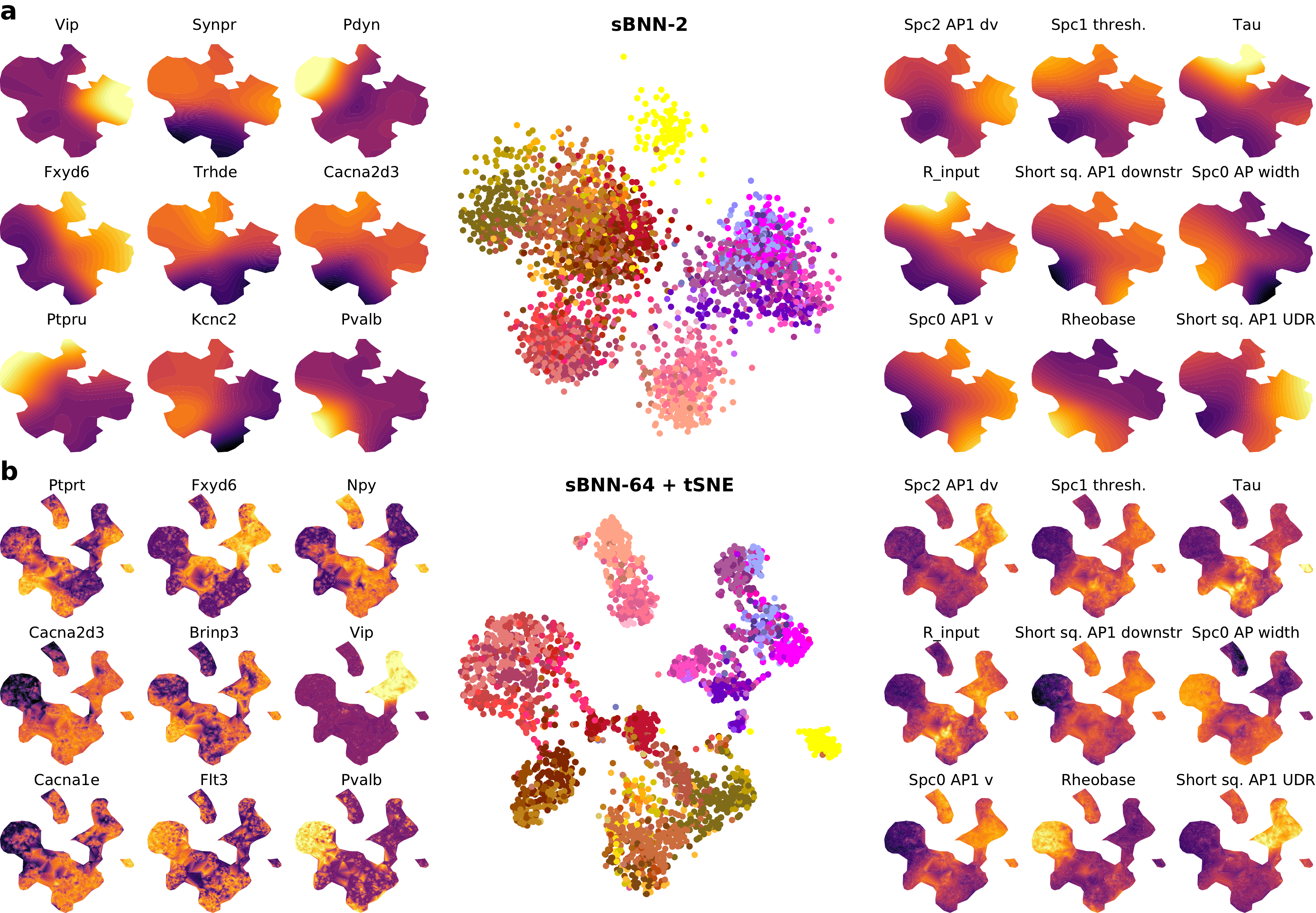}
\caption{The V1 dataset\cite{gouwens2020toward}, $n=3395$ cells are shown. Same format as in Figure~\ref{fig:scala_full_dataset}.}
\label{fig:gouwens_full_dataset}
\end{figure*}

The two-dimensional latent bottleneck representation of the same final (trained on the entire dataset) sBNN-2 model can be directly visualized in compact form together with its relationship to gene expression and electrophysiological properties (Figures~\ref{fig:scala_full_dataset}a,  \ref{fig:gouwens_full_dataset}a). In contrast, for the sBNN-64 model, we performed t-SNE to create a 2D embedding of the latent space for the purpose of visualization (Figures~\ref{fig:scala_full_dataset}b, \ref{fig:gouwens_full_dataset}b). 

In both cases, major known groups of neurons (excitatory neurons and several families of interneurons) were well separated in the latent spaces for both datasets. Overlaying the selected gene expression reconstruction from the autoencoder model predictions on the latent space  (Figures~\ref{fig:scala_full_dataset}--\ref{fig:gouwens_full_dataset}, left) rediscovered well-known marker genes for these major cell families. For example, \textit{Pvalb}, and \textit{Vip} genes were selected by both models (sBNN-2 and sBNN-64) for both datasets (M1 and V1), and clearly distinguished specific groups of neurons in the visualization, in agreement with the literature \cite{tasic2018shared, trembley2016GABA}. 

Likewise, overlaying model predictions of electrophysiological features on this embedding (Figures~\ref{fig:scala_full_dataset}--\ref{fig:gouwens_full_dataset}, right) highlighted key known properties of these major cell groups. For example, in the \textit{Pvalb} family all transcriptomic types (different shades of red) were fast-spiking neurons with high firing rates and narrow action potentials, whereas e.g. in the \textit{Vip} family different transcriptomic types (shades of purple) were all characterized by high input resistance, low rheobase and a propensity for rebound firing. Importantly, the sBNN extracted these major cell families, their transcriptomic signatures and their defining electrophysiological properties from the data without any biological prior knowledge being incorporated into the model. 

Our sBNNs also allow for meaningful biological interpretations beyond the level of major cell families: for instance for \textit{Vip} neurons (Figure~\ref{fig:scala_full_dataset}a,b purple colors) we found a clear separation between different t-types in the latent space, with differences in several predicted electrophysiological properties such as the membrane time constant and rebound, and predicted gene expression level such as \textit{Fxyd6} (Figure~\ref{fig:scala_full_dataset_zoomed_in}) \cite{scala2020phenotypic} in line with experimental values.

For comparison, we provide the same kind of visualization for the linear sRRR-2 and sRRR-full model (Supplementary Figure~\ref{fig:scala_full_dataset_linear}). In the first two latent dimensions for the M1 dataset, \textit{Vip} and \textit{Sst} populations were overlapping, despite having very different gene expression and firing properties. Indeed, only the third sRRR component separated these two families of neurons \cite{scala2020phenotypic}. As sRRR is a linear model, all model predictions were linear in the embedding space (Supplementary Figure~\ref{fig:scala_full_dataset_linear}a, left and right), in contrast to the non-linear surfaces that model predictions formed over the latent space of the sBNN-2 model (Figure~\ref{fig:scala_full_dataset}a, left and right).

\subsection*{Ion channel genes}

Interestingly, selected genes beyond the known cell class markers often had direct and interpretable relations to ion channel dynamics and the predicted electrophysiological properties. For example, \textit{Kcnc2}, selected by the sBNN-64 for the M1 dataset, encodes a potassium voltage-gated channel and \textit{Slc6a1}, selected by the sBNN-2, encodes a sodium- and chloride-dependent GABA transporter. The sBNN-based visualizations of the M1 dataset revealed that \textit{Kcnc2} and \textit{Slc6a1} expression was higher in the fast spiking \textit{Pvalb} interneurons, providing a causal hypothesis, which could be experimentally tested (Figure~\ref{fig:scala_full_dataset}) \cite{erisir1999fastspikingcause, karsten2007differential, Jiang2016epilepsy}. Likewise, \textit{Cacna1e}, selected in both the M1 and V1 datasets, encodes a voltage-sensitive calcium channel, which was more highly expressed in cell families with high afterhyperpolarization such as excitatory neurons or \textit{Vip} interneurons \cite{zaman2011cacna1eAHP}.

We did an additional experiment to see if electrophysiological variability could be predicted by using the ion channel genes alone. We constrained our sBNN-2 and sBNN-64 model to use a set of 423 ion channel genes, and observed only a modest drop in cross-validated multivariate $R^2$ performance (from $0.39$ to $0.35$) for the sBNN-2 model. We also found that restricting the gene set strongly improved gene selection stability for the sBNN-2 model as $13$ instead of two genes were selected in all training repetitions, and even $19$ instead of $16$ genes for the sBNN-64 model, compared to our initial experiment (see Methods). Many of the selected genes, regarding for instance sBNN-2, corresponded to potassium channels (e.g. \textit{Kcnh4, Kcnh7, Kcnc1, Kcnc2}) and calcium voltage-gated channels (\textit{Cacna2d1, Cacna2d2}, and \textit{Cacna1e}) (Supplementary Figure~\ref{fig:scala_latent_ion_channel_genes}).

\subsection*{Beyond Patch-seq}

\begin{figure*}[ht]
\centering
\includegraphics[width=\linewidth]{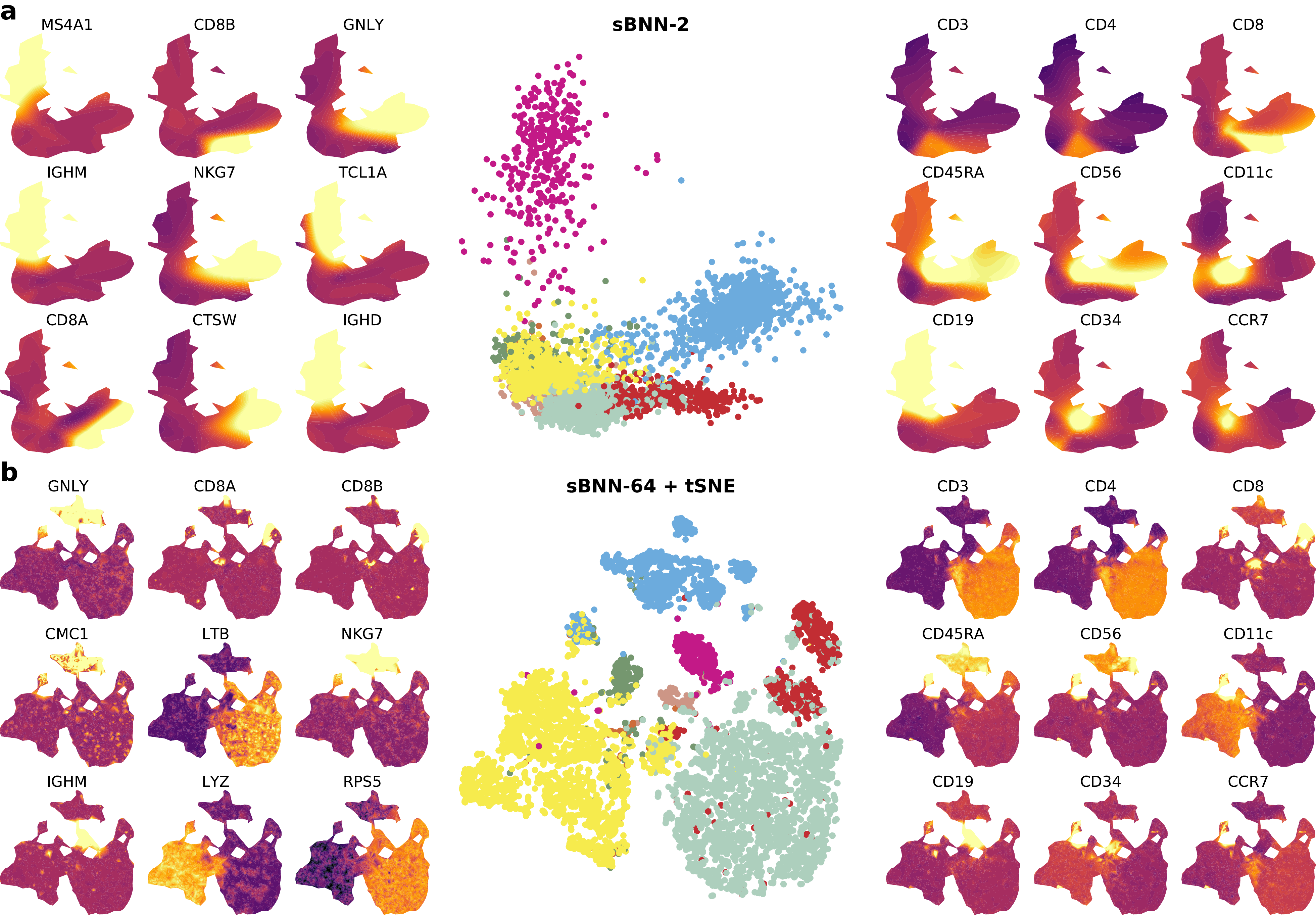}
\caption{The CITE-seq dataset\cite{stoeckius2017cite-seq}, $n=7652$ cells are shown. Same format as in Figure~\ref{fig:scala_full_dataset}. Cell identities and cluster colors were taken from the original publication. Red: CD8 T cells; green: CD16 monocytes; yellow: CD14 monocytes; blue: natural killer cells; aquamarine: CD4 T cells; magenta: B cells; brown: precursors. Colorscale in all overlay subplots goes from $-1$ (dark blue) to $1$ (bright yellow).}
\label{fig:stoeckius_latent}
\end{figure*}

To show that sBNN framework can be used for analysis of paired data other than Patch-seq, we applied it to a CITE-seq dataset \cite{stoeckius2017cite-seq} providing a detailed multimodal characterization of cord blood mononuclear cells and in which transcriptomic and protein marker measurements are combined in the same cells. The performance of sBNN-2 (cross-validated $R^2=0.38$, Table~\ref{table:sBNN_vs_sRRR_R2_scores}) not only far outperformed the rank-2 linear model ($R^2=0.23$) but also the full-rank linear model ($R^2=0.35$), demonstrating again the usefulness of nonlinear approaches for single-cell analysis (Supplementary Figure~\ref{fig:stoeckius_training_curves}). Our method produced biologically meaningful latent visualizations (Figure~\ref{fig:stoeckius_latent}), without any modifications to the training regime. For instance, the model correctly predicted high \textit{CD4} and \textit{CD8} protein expression in corresponding cell types (Figure \ref{fig:stoeckius_latent}), while marker genes like \textit{CD8A} and \textit{CD8B} were selected into the model after pruning.

\section*{Discussion}

Patch-seq has emerged as an important technique for understanding cellular diversity within and between cell types by allowing to characterize relationships between different data modalities \cite{cadwell2017multimodal, Camunas-Soler2019pancreas, scala2020phenotypic, scala2019layer, TripathyKobakPatchSeq}. Resulting multimodal data create a need for machine learning and data science to come up with algorithms and frameworks suitable to study the relationship between transcriptomic, electrophysiological, and morphological properties \cite{scala2020phenotypic, scala2019layer, gouwens2020toward, Bugeon2021geneticaxis}.

Here we build upon our previously published linear approach \cite{kobak2019sparse} and introduce a non-linear framework based on deep neural networks. Our approach compresses one view of the data (transcriptome) into a latent representation from which we  predict other views of the data. In addition, the latent space allows us to directly visualize the multimodal data for data exploration. We showed that our sBNN framework with group lasso penalty can detect marker and ion channel genes that directly relate to well-known neuronal families and specific electrophysiological behaviour, without relying on prior biological knowledge. This allows to hypothesize which genes could be responsible for specific phenotypic behaviour, which is highly relevant for guiding further experimental work. In addition to demonstrating its efficacy for Patch-seq, we also show that our sBNN framework can be applied to other types of data such as CITE-seq, where it can accurately predict proteomic measurements from gene expression levels.

We studied two versions of the sBNN framework: sBNN-2 with a two-dimensional bottleneck, and sBNN-64 with a 64-dimensional bottleneck. We found that sBNN-64 outperformed the sBNN-2 model in terms of predictive performance (Table~ \ref{table:sBNN_vs_sRRR_R2_scores}) and showed reduced variability in predicting individual features across cross-validation folds (Figure~\ref{fig:ind_ephys_scores}). Moreover, sBNN-64 showed greater stability in its gene selection (Figure~\ref{fig:gene_histograms}). On the other hand, the two-dimensional latent bottleneck representation of the sBNN-2 model allows for direct visualization (Figure~\ref{fig:scala_full_dataset}a), whereas sBNN-64 requires an additional dimensionality reduction step such as t-SNE (Figure \ref{fig:scala_full_dataset}b).

Related to our work, a coupled autoencoder framework has been recently suggested for Patch-seq data visualization \cite{gala2019coupled,gala2020consistent}. There, two different autoencoder networks are trained on the transcriptomic and on the electrophysiological data but are constrained to have similar latent spaces. Our sBNN framework is different in two ways: (i) we directly predict electrophysiological properties from gene expression levels, which is a biologically motivated choice that allows us to train a single neural network; and (ii) we use a group lasso penalty to perform a biologically meaningful gene selection.

In conclusion, we demonstrated that the sBNN framework yields compact, non-linear, and biologically interpretable visualizations of Patch-seq data and beyond. Our method can detect specific marker and transmembrane genes that are directly related to electrophysiological and cell-surface proteomic properties, allowing to form testable hypotheses. We believe that our framework can also be applied to other kinds of multimodal datasets in single-cell biology and beyond.

\newcommand{\X}{\mathbf X}
\newcommand{\Y}{\mathbf Y}
\newcommand{\B}{\mathbf B}
\newcommand{\Bhat}{\mathbf{\hat B}}
\newcommand{\w}{\mathbf w}
\newcommand{\W}{\mathbf W}
\newcommand{\V}{\mathbf V}
\newcommand{\I}{\mathbf I}
\renewcommand{\v}{\mathbf v}
\renewcommand{\L}{\mathcal L}
\newcommand{\tr}{\operatorname{tr}}

\vspace{4em}
\noindent\rule{\columnwidth}{.5pt}
\section{Methods}
\phantomsection

\subsection{Code availability}

The code in Python is available at \url{https://github.com/berenslab/sBNN}.

\subsection{Data sources and preprocessing}

The M1 data were downloaded from \url{https://github.com/berenslab/mini-atlas} in form of transcriptomic read counts and extracted electrophysiological features. We used the same standard \cite{luecken2019current} preprocessing steps as in the original publication \cite{scala2020phenotypic} ending up with 16 electrophysiological features, 1213 cells that passed the quality control and were assigned a transcriptomic type, and 1000 genes that were most variable across this subset of cells. We divided the gene counts for each cell by the cell sum over all genes (sequencing depth) and multiplied the result by the median sequencing depth size across all cells. We then log-transformed the data using $\log_2(x+1)$ transformation. Finally, all features in $\X$ and $\Y$ were standardized to have zero mean and unit variance. See Supplementary Figure~\ref{fig:scala_tsne} for separate visualizations of both modalities using t-SNE \cite{maaten2008visualizing} (see Figure~\ref{fig:scala_full_dataset}b for t-SNE embedding of the sBNN-64 latent space).

The V1 data were downloaded from \url{https://github.com/AllenInstitute/coupledAE-patchseq} also in form of transcriptomic read counts and extracted electrophysiological features. This dataset \cite{gouwens2020toward} already contained normalized and log-transformed gene expression values and normalized electrophysiological feature values \cite{gala2020consistent} but for consistency we further $z$-scored the log-expression of all genes, excluded neurons that showed undefined electrophysiological feature values, $z$-scored every electrophysiological feature and excluded some of them that showed particularly high correlation with other features.

The CITE-seq data \cite{stoeckius2017cite-seq} were downloaded from \url{https://www.ncbi.nlm.nih.gov/geo/query/acc.cgi?acc=GSE100866} in form of transcriptomic read counts and epitomic measurements. Similarly to M1 and V1 datasets, we kept 7652 cells that were assigned a cell type specific to the immune system (excluding erythrocytes, dendritic cells and unassigned cells), all 13 epitomic features, selected 1000 most variable genes, divided the gene counts for each cell by the cell sum over all genes (sequencing depth) and multiplied the result by the median sequencing depth size across all cells. We then log-transformed the data using $\log_2(x+1)$ transformation. Finally, all features in $\X$ and $\Y$ were standardized to have zero mean and unit variance.

In the end, the $\X$ and $\Y$ data matrices had shapes $1213\times 1000$, $1213\times 16$ respectively for the M1 dataset, $3395\times 1252$ and $3395\times 55$ for the M1 dataset and $7652\times 1000$ and $7652\times 13$ for the CITE-seq dataset.

\subsection{List of electrophysiological features}

For the M1 dataset we used the following electrophysiological features: action potential (AP) amplitude, AP amplitude adaptation, AP amplitude coefficient of variation, AP threshold, AP width, afterhyperpolarization, adaptation index, interspike interval coefficient of variation,
input resistance, max AP number, membrane time constant, rebound, resting membrane potential, rheobase, hyperpolarization sag, and the upstroke-downstroke ratio. See the original paper \cite{scala2020phenotypic} for exact definitions. For the V1 dataset we refer to the original paper \cite{gouwens2020toward}.

\subsection{List of ion channel genes}
We downloaded the names of $1992$ ion channel genes from \url{http://www.informatics.jax.org/go/term/GO:0005216} and used $423$ of them that were also contained in the M1 dataset.

\subsection{Reduced-rank regression}

This paper builds on previous work suggesting to use sRRR \cite{chen2012sparse} for visualization of paired datasets in genetics and neuroscience \cite{kobak2019sparse}. In general, RRR \cite{izenman1975reduced, velu2013multivariate} is linear regression with a low-rank constraint on the weight matrix. Its loss function can be written as
\begin{equation}
    \L = \lVert \Y - \X\W\V^\top \rVert^2,  
\end{equation}
where $\X$ and $\Y$ are centered predictor and response data matrices, and $\W$ and $\V$ both have $r$ columns, resulting in a rank-$r$ weight matrix $\W\V^\top$. Using autoencoder terminology, $\W$ and $\V$ are an encoder and a decoder, respectively. Without loss of generality, it is convenient to require that $\V$ has orthogonal columns, i.e. $\V^\top\V=\I$. RRR can be directly solved by singular value decomposition (SVD), as can be seen by decomposing its loss into the unconstrained least squares loss and the low-rank loss \cite{kobak2019sparse}. Regularization with a ridge penalty still allows for the direct solution using SVD.

Sparse RRR \cite{chen2012sparse} imposes a group lasso penalty \cite{yuan2006model}  on the encoder coefficients: 
\begin{equation}
    \L = \lVert \Y - \X\W\V^\top \rVert^2 + \lambda \sum_{i=1}^p\lVert \mathbf W_{i\cdot}\rVert_2.   
\end{equation}
Here $\sum_{i=1}^p\lVert \mathbf W_{i\cdot}\rVert_2 = \sum_{i=1}^p\sqrt{\sum_{j=1}^r W_{ij}^2}$ is an $\ell_1$ norm of $\ell_2$ row norms. The group lasso encourages the encoder $\W$ to have sparse rows (instead of sparse individual elements), performing feature selection. It is possible to add a further ridge penalty, arriving at RRR with elastic net regularization \cite{kobak2019sparse}. SRRR is a bi-convex problem and can be solved by alternating steps: for fixed $\V$, the optimal decoder can be found using \texttt{glmnet} \cite{friedman2010regularization}, while for fixed $\W$, the problem reduces to a Procrustes problem and can be solved using SVD \cite{yuan2006model, kobak2019sparse}.

Lasso and elastic net penalties can sometimes lead to over-shrinkage \cite{efron2004least, zou2005regularization, meinshausen2007relaxed} and a two-stage procedure that re-fits the model using selected predictors can lead to a higher predictive performance \cite{efron2004least, meinshausen2007relaxed, mol2009regularized, kobak2019sparse}. In particular,  `relaxed lasso' \cite{meinshausen2007relaxed} fits the lasso model with some regularization parameter $\lambda$ and then re-fits it using only selected predictors  (i.e. predictors with non-zero coefficients) and another, usually smaller, regularization parameter $\lambda_2$. Here we used a `relaxed elastic net' variant that uses ridge penalty with the same regularization strength $\lambda$ for the second-stage fit \cite{kobak2019sparse}. This way there is only one hyperparameter to tune.

SRRR models in this work either had rank $r=2$ or rank $r=16$ for the M1 and V1 dataset where rank $r=16$ corresponded to the full dimensionality of $\Y$ in the M1 dataset. Regarding the CITE-seq dataset, the full rank sRRR model corresponded to rank $r=13$.

\subsection{Bottleneck neural network}

In our sparse bottleneck neural network (sBNN) architecture, the encoder and the decoder have two hidden nonlinear layers each (we use 512 and 128 units for the encoder, and 128 and 512 units for the decoder) and either 64 or 2 units in the bottleneck layer itself to which we gave the names sBNN-64 and sBNN-2 respectively. We used exponential linear units for these hidden layers, but kept the bottleneck layer and the output layer linear, to match the RRR model as closely as possible (Figure~\ref{fig:schema}) and also because the response variables can take values in $\mathbb R$ and are not necessarily constrained to be positive or to lie between 0 and 1. We used the mean squared error (MSE) loss function. The model was implemented using the \texttt{Keras} (version 2.7.0) and \texttt{TensorFlow} (version 2.7.0) libraries in Python.

We used a constant $\ell_2$ penalty of $10^{-10}$ on all layers (including bias units in all layers apart from the readout layer). We found that the model performance was unaffected in a broad range of the penalty values (from $10^{-12}$ to $10^{-4}$).

\subsection{Latent visualizations}
The linear sRRR-2 and nonlinear sBNN-2 frameworks naturally allow for direct two-dimensional visualizations and interpretation of the data as the bottleneck in both scenarios is already two-dimensional. In the case of our sRRR-full and sBNN-16 frameworks an additional step is required. Here, we used the \texttt{openTSNE} implementation with default parameters (\url{https://github.com/pavlin-policar/openTSNE}) to embed the latent space with t-sne to a final two-dimensional representation.

In order to overlay latent space visualizations with ephys and gene expression level predictions we first computed model predictions for all cells with our trained neural network and then used the \texttt{matplotlib.pyplot\linebreak .tricontourf} function to plot the contours on the triangulation.
Colorscale in all overlay subplots goes from $-1$ (dark blue) to $1$ (bright yellow).

\subsection{Group lasso and pruning for feature selection}

To enforce sparsity, we imposed the same group lasso penalty as above on the first layer weights. This was implemented as a custom penalty in \texttt{Keras}. Note that there exist other approaches to perform feature selection \cite{yamada2014featurewisekernelizedlasso, yamada2020gates}, but we chose group lasso for simplicity and for better correspondence to sparse reduced-rank regression.

The strength of the lasso penalty played an important role (Supplementary Figure~\ref{fig:different_lasso}). For low penalties such as 0.0001, the $\ell_2$ weight norms for different input units before pruning were all of similar size (Supplementary Figure~\ref{fig:different_lasso}b), leading to arbitrary genes being selected after pruning and bad performance afterwards (Supplementary Figure~\ref{fig:different_lasso}a). Strong penalties (0.01--1.0) led to rapidly decaying weight norms (Supplementary Figure~\ref{fig:different_lasso}b), similar sets of genes being selected, and similar final performance (Supplementary Figure~\ref{fig:different_lasso}a), with 0.1 penalty achieving the highest performance. We used the 0.1 penalty for all models shown in Figures~\ref{fig:scala_full_dataset}-\ref{fig:stoeckius_latent}.

Linear models with lasso penalty allow for exact solutions that converge to having entries which are exactly zero \cite{friedman2010regularization}. This is not the case for deep learning models, where gradient descent will generally fluctuate around solutions with many small but non-zero elements. Our strategy to achieve a genuinely sparse model was to (i) impose a strong lasso penalty and train the network until convergence; then (ii) prune the input layer by only selecting a pre-specified number of input units with the highest $\ell_2$ row norms in the first layer, and deleting all the other input units; and finally, (iii) fine-tune the resulting model with lasso penalty set to zero. This procedure mimics the `relaxed' elastic net procedure described above: the lasso-regularized model is pruned and then fine-tuned without a lasso penalty.

We observed a curious effect when imposing the group lasso penalty: while the training and test performance were initially both improving, after several epochs they both deteriorated rapidly (Figure~\ref{fig:training_curves}, Figure~\ref{fig:training_curves_tr}b,c). This was not due to overfitting because the training performance became worse as well (Figure~\ref{fig:training_curves_tr}b,c). Instead, the shape of the performance curves suggests that the lasso penalty `began kicking in', bringing the penalized MSE loss down while the unpenalized MSE loss went up. After this initial rapid decrease, the test performance slowly improved again with further training. We speculate that this behavior suggests a phase transition in the gradient descent dynamics: in the earlier $\sim$10 epochs mainly the MSE (or cross-entropy) term was being optimized, while in the next $\sim$10 epochs it was mainly the lasso penalty. We observed qualitatively the same phase transition when using the normal (element-wise) lasso penalty as implemented in \texttt{Keras}, so it was not specific to the group lasso penalty we used. 

\subsection*{Training}

We used the Adam optimizer \cite{Adam} with learning rate $0.0001$, which we found to perform better on our dataset (possibly due to the relatively low sample sizes, see below). The default value $0.001$ sometimes led to very noisy convergence, especially during pre-training (see below). All layers were initialized using the \texttt{glorot\_uniform} initializer \cite{glorot2010understanding}, which is default in \texttt{Keras}. The batch size for stochastic gradient descent was set to $32$. It took $\sim$1 minute to train any of these models for the entire 250 epochs on NVIDIA Titan Xp.

\subsection{Pre-training}

We performed pre-training in a classification setting with a categorical cross-entropy loss. This approach was inspired by the usual practice for image-based regression tasks to start with a convolutional neural network with weights trained for classification on ImageNet and to fine-tune them on the task at hand \cite{lathuiliere2019comprehensive}. Also, it has been shown that transferring the weights from an initial task A to a target task B with subsequent fine-tuning can improve the performance on task B compared to random weight initialization \cite{transferlearning}.

We used $k$-means clustering to cluster the whole dataset of $n$ points into $K$ clusters based on the values of the response variables. We then replaced the output layer of our network with a $K$-element softmax and trained the network to predict the cluster identity using the cross-entropy loss. This can be seen as a coarse-grained prediction because the actual $\Y$ values were replaced by the cluster identities. We used $K = 20$ for all experiments.

We held out 40\% of the training set as the validation set in order to perform early stopping and choose the training epoch that had the lowest unpenalized cross-entropy validation loss (during 50 epochs). We used the weights obtained at that epoch as a starting point for subsequent training with the original MSE loss. The output layer (that was not pre-trained) was initialized using the default \texttt{glorot\_uniform} initializer.

A common advice in transfer learning is to hold the bottom layers frozen after the transfer (as those are the most generalizable \cite{transferlearning}), train the upper layers first and then unfreeze all layers for fine-tuning \cite{howard2018universal}. We tried this procedure and held the bottom two layers frozen for the first 50 epochs, and then trained the network with all layers unfrozen for another 50 epochs. In our experiments, the performance was very similar for any number of frozen layers from two to five. We found it advantageous to reduce the learning rate after unfreezing to $0.00005$ (the same learning rate was also used after pruning).

\subsection{Model selection}

The sample sizes of the Patch-seq datasets used in this work were very moderate by deep learning standards ($n=1213$ for the M1 dataset and $n=3395$ for the V1 dataset). Therefore, it was not possible to have a large test set, and evaluating performance on a small test set would yield a noisy estimate. To mitigate this problem, we used 10-fold cross-validation (CV). All models were fit 10 times on training sets containing 90\% of the data and evaluated on the remaining 10\%. All performance curves shown in the paper are averages over the 10 folds. The cross-validation folds were the same for all models. 

We used $R^2$ as the performance metric for the regression models. We computed the multivariate test-set $R^2$ score as 
\begin{equation}
   R_\mathrm{test}^2 = 1- \frac{\lVert \Y_\mathrm{test}-f(\X_\mathrm{test})\rVert^2}{\lVert \Y_\mathrm{test}\rVert^2},
\end{equation}
where $f(\cdot)$ is the sRRR or the sBNN output and $\X_\mathrm{test}$ and $\Y_\mathrm{test}$ were centered using the corresponding training-set means. $R_\mathrm{train}^2$ was defined analogously.  For linear models, this gives the standard expression for the $R^2$ (coefficient~of~determination). Finally, we defined test-set $R^2$ score for an individual electrophysiological feature $i$ similarly as
\begin{equation}
   R_i^2 = 1- \frac{\lVert [\Y_\mathrm{test}]_i-f_i(\X_\mathrm{test})\rVert^2}{\lVert [\Y_\mathrm{test}]_i\rVert^2}.
\end{equation}

\subsection{Training on the entire dataset}

We chose the best hyperparameters and optimisation approach based on the cross-validation results, but for the purposes of visualization, we re-trained the chosen model using the entire dataset. We then passed the whole dataset through the model, to obtain the bottleneck representation, which we used for visualizations. We used the same re-trained model to analyse which features were selected into the sparse model.

\subsection{Model stability}
\label{model_stability}

Retraining the model may yield a different set of selected genes, due to randomness in the initialization and stochastic gradient descent optimization. As always in sparse models, there is a trade-off between sparsity and model stability \cite{xu2011sparse}. Using the entire M1 dataset, we trained the the sBNN-2 model nine more times and found that for each pair of runs, on average 13 genes were selected by both of them. 2 genes were selected 10 times out of 10: \textit{Pvalb} and \textit{Gad1} (Figure~\ref{fig:gene_histograms}). Latent visualizations corresponding to different random initializations all looked qualitatively similar and suggested the same biological interpretations (Supplementary Figure~\ref{fig:instability}). Likewise, for the V1 dataset and sBNN-2 model, on average 14 genes were selected for each pair of runs, and 6 genes were selected 10 times out of 10: \textit{Pvalb, Vip, Zfp536, Egfem1, Lamp5} and \textit{Pdyn}.

Increasing the dimensionality of the bottleneck layer to 64 increased model stability: in the V1 dataset, 16 genes were selected 10 times out of 10, and, for each pair of runs, on average 21 genes were selected by both of them (Figure~\ref{fig:gene_histograms}).

When training the sBNN-2 model using ion channel genes only (Supplementary Figure~\ref{fig:scala_latent_ion_channel_genes}), we found 13 genes selected 10 times out of 10.

\bibliography{main}

\section*{Acknowledgements}

We thank Federico Scala, Andreas S. Tolias and Rickard Sandberg for discussion. This work was funded by the German Ministry of Education and Research (FKZ 01GQ1601, 01IS18039A), the German Research Foundation (EXC 2064 project number 390727645, BE5601/4-1) and the National Institute Of Mental Health of the National Institutes of Health under Award Number U19MH114830. The content is solely the responsibility of the authors and does not necessarily represent the official views of the National Institutes of Health.

\section*{Author contributions statement}

YB, PB, and DK conceptualized the project. YB performed computational experiments. YB, DK and PB wrote the paper.

\section*{Declaration of interests}
The authors declare no competing interests.

\onecolumn
\section*{Supplementary figures}

\renewcommand{\thefigure}{S\arabic{figure}}
\setcounter{figure}{0}

\begin{figure}[ht]
\begin{center}
\centerline{\includegraphics[width=\textwidth]{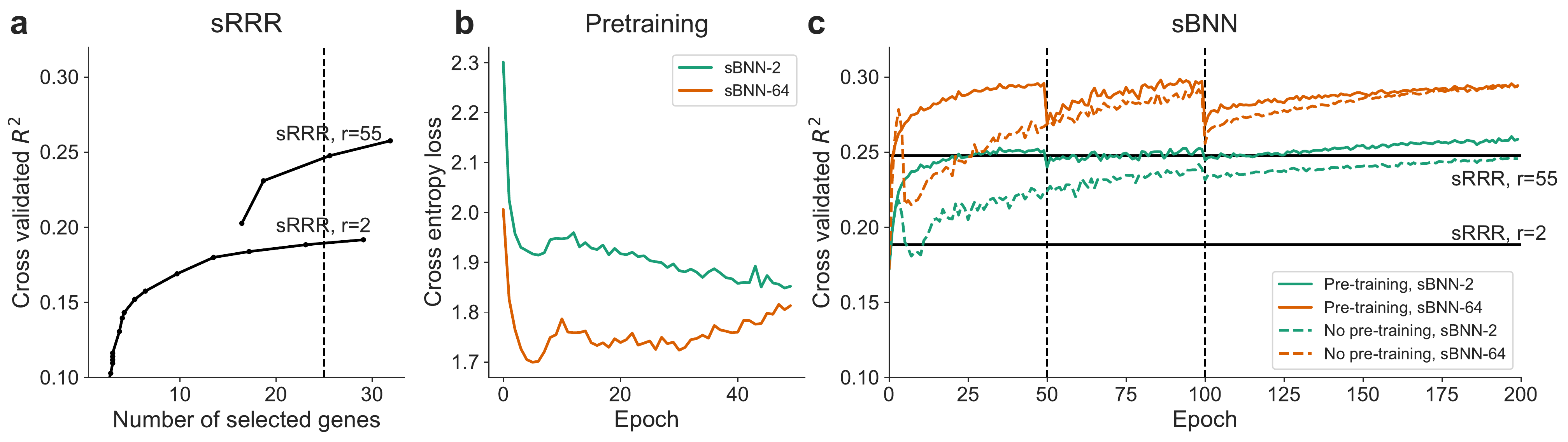}}
\caption{V1 dataset. \textbf{a} sRRR rank-2 and full-rank cross-validated $R^2$ scores. Vertical line corresponds to regularizing the linear model such that 25 genes are selected. \textbf{(b)} Pretraining sBNN-2 and sBNN-64 validation cross-entropy scores. \textbf{c} Linear and nonlinear sparse models, trained on the V1 dataset. Same format as in Figure~\ref{fig:training_curves}.}
\label{fig:training_curves_Allen}
\end{center}
\end{figure}

\begin{figure}[ht]
\begin{center}
\centerline{\includegraphics[width=\textwidth]{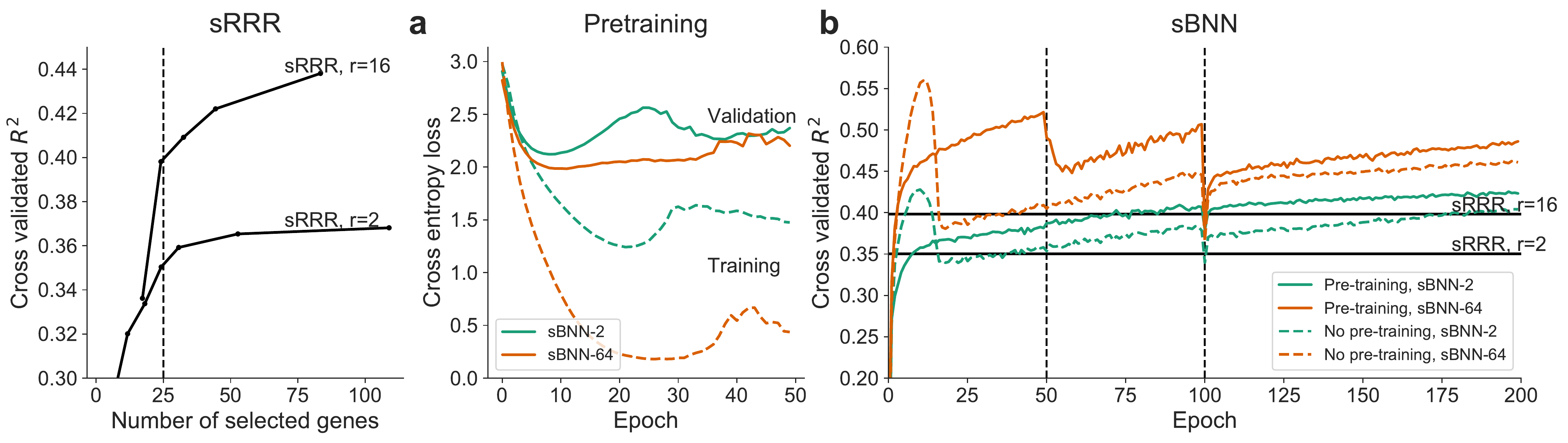}}
\caption{M1 dataset. \textbf{a} sRRR rank-2 and full-rank cross-validated $R^2$ scores. Vertical line corresponds to regularizing the linear model such that 25 genes are selected. \textbf{(b)} Pretraining sBNN-2 and sBNN-64 validation and training set cross-entropy scores. \textbf{(c)} Same format as in Figure~\ref{fig:training_curves}, but for training set performances.}
\label{fig:training_curves_tr}
\end{center}
\end{figure}

\begin{figure}[ht]
\centering
\includegraphics[width=\linewidth]{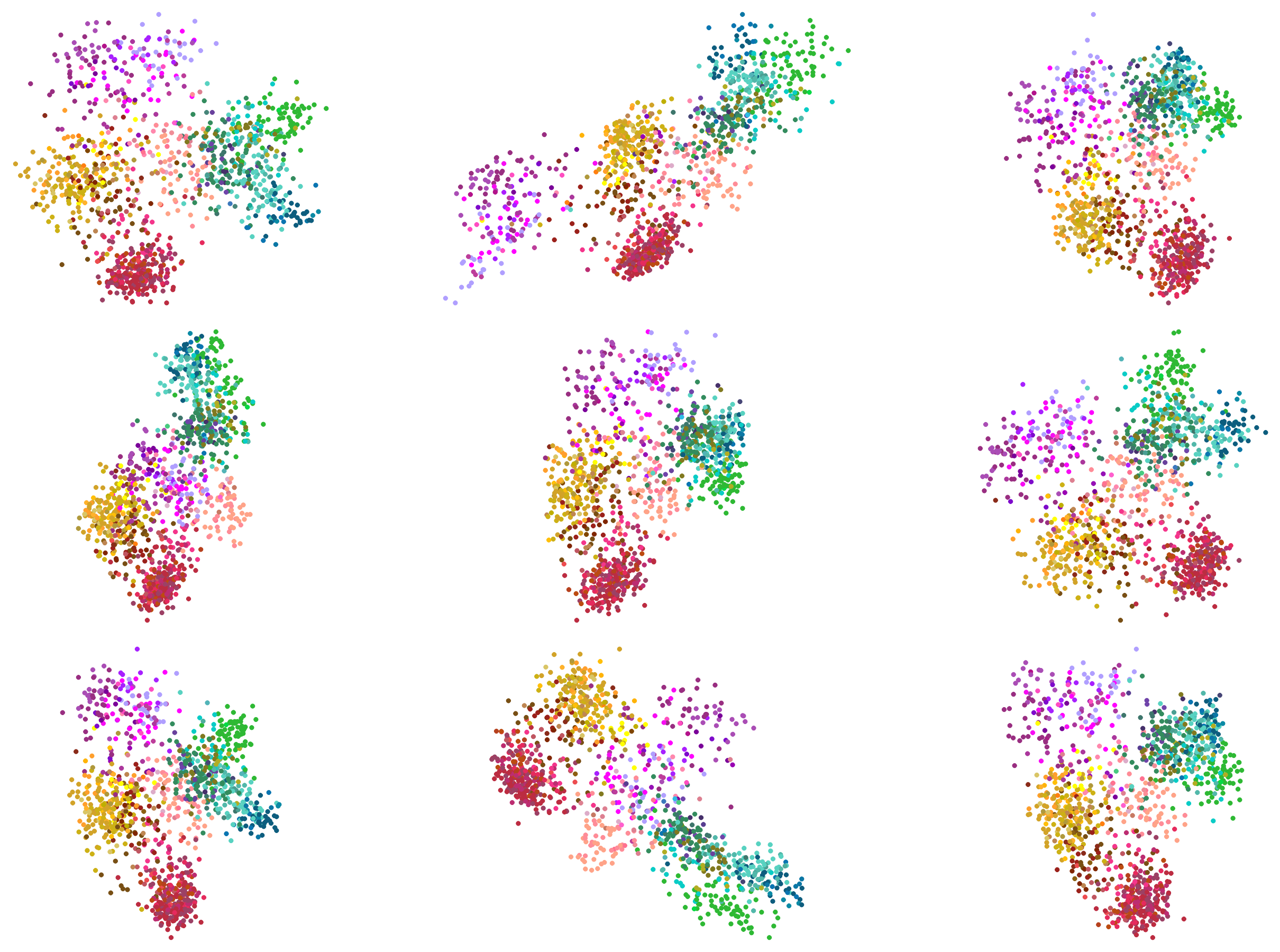}
\caption{Latent representations corresponding to different random initializations (9 out of 10 shown).}
\label{fig:instability}
\end{figure}

\begin{figure}[ht]
\centering
\includegraphics[width=\linewidth]{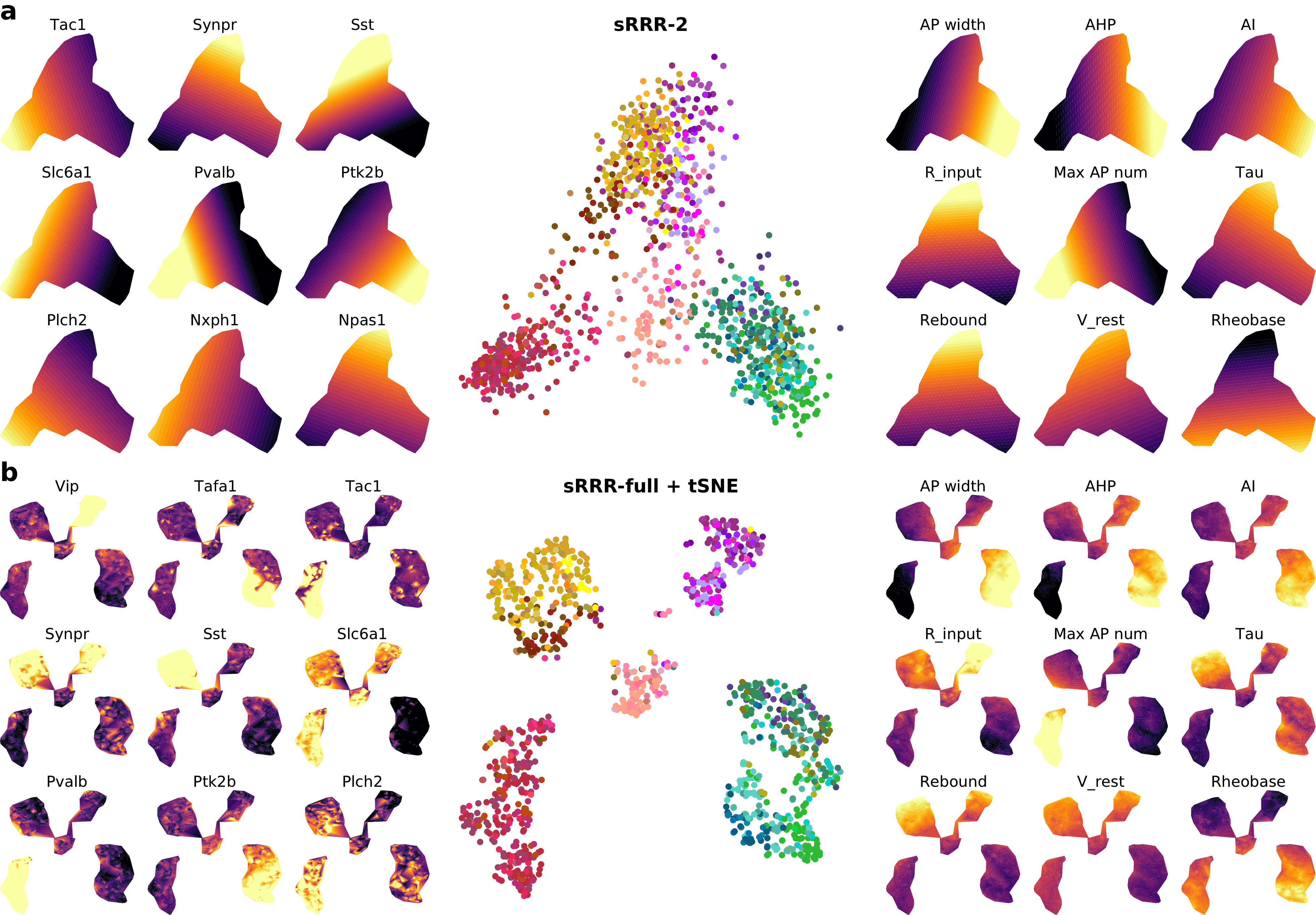}
\caption{The latent spaces of the relaxed sRRR models trained on the M1 dataset. Colors correspond to transcriptomic types and are taken from the original publications. Red: Pvalb interneurons; yellow: Sst interneurons;
purple: Vip interneurons; salmon: Lamp5 interneurons; green/blue: excitatory neurons. Colorscale in all overlay subplots goes from $-1$ (dark blue) to $1$ (bright yellow). Same format as
in Figure~\ref{fig:scala_full_dataset}. \textbf{a} Full-rank sRRR. \textbf{b} Rank-2 sRRR.}
\label{fig:scala_full_dataset_linear}
\end{figure}

\begin{figure}[ht]
\centering
\includegraphics[width=\linewidth]{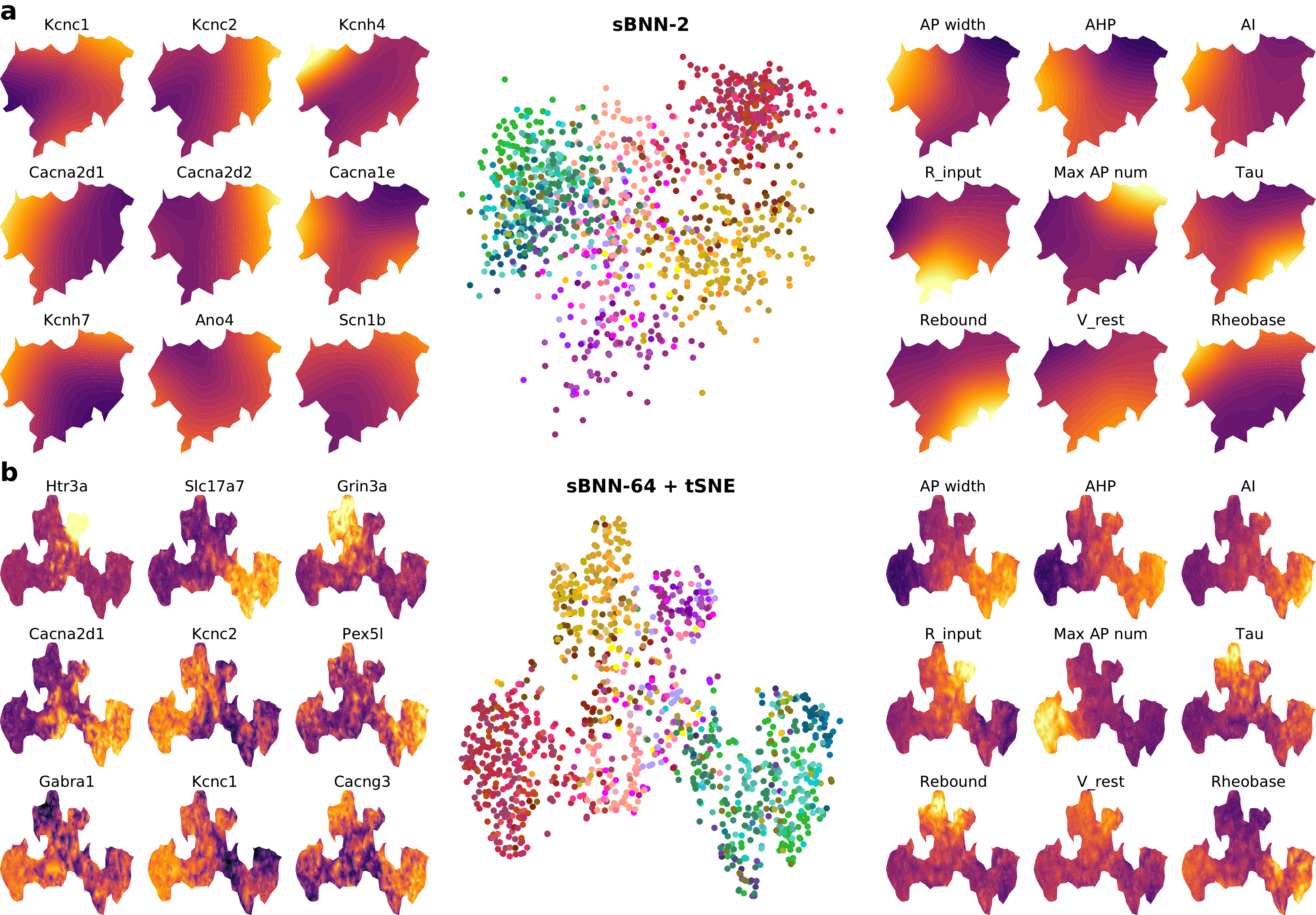}
\caption{The M1 dataset. Same format as in Figure~\ref{fig:scala_full_dataset}, but using $n=423$ ion channel genes only. The list of ion channel genes was taken from \url{http://www.informatics.jax.org/go/term/GO:0005216} (see Methods).}
\label{fig:scala_latent_ion_channel_genes}
\end{figure}

\begin{figure}[ht]
\centering
\includegraphics[width=\textwidth]{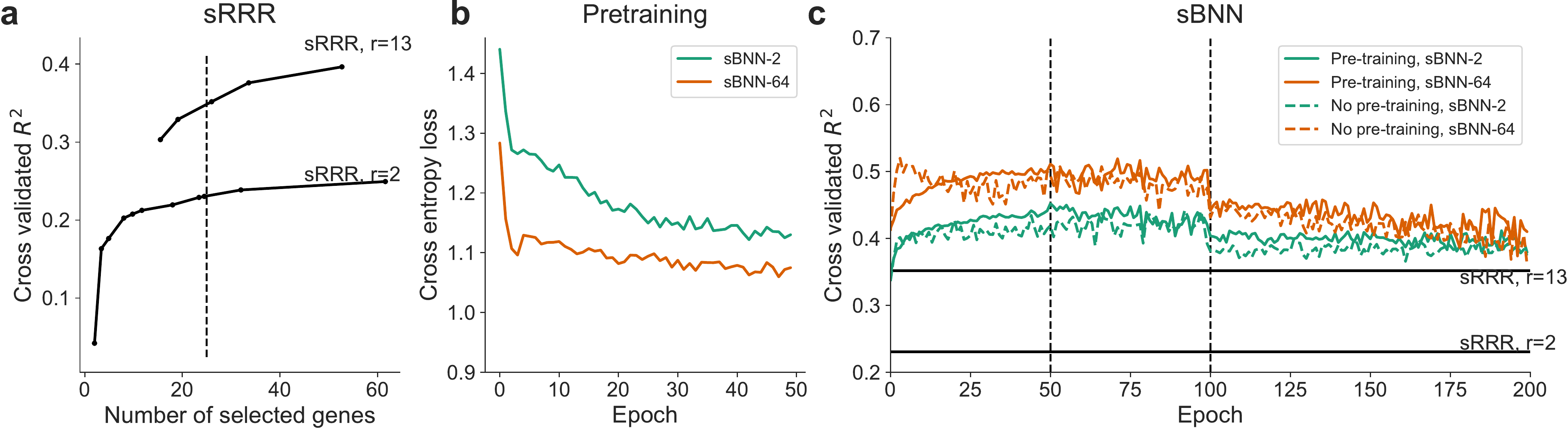}
\caption{Linear and nonlinear sparse models, trained on the CITE-seq dataset. Same format as in Supplementary Figure~\ref{fig:training_curves_Allen}.}
\label{fig:stoeckius_training_curves}
\end{figure}

\begin{figure}[ht]
\centering
\includegraphics[width=\linewidth]{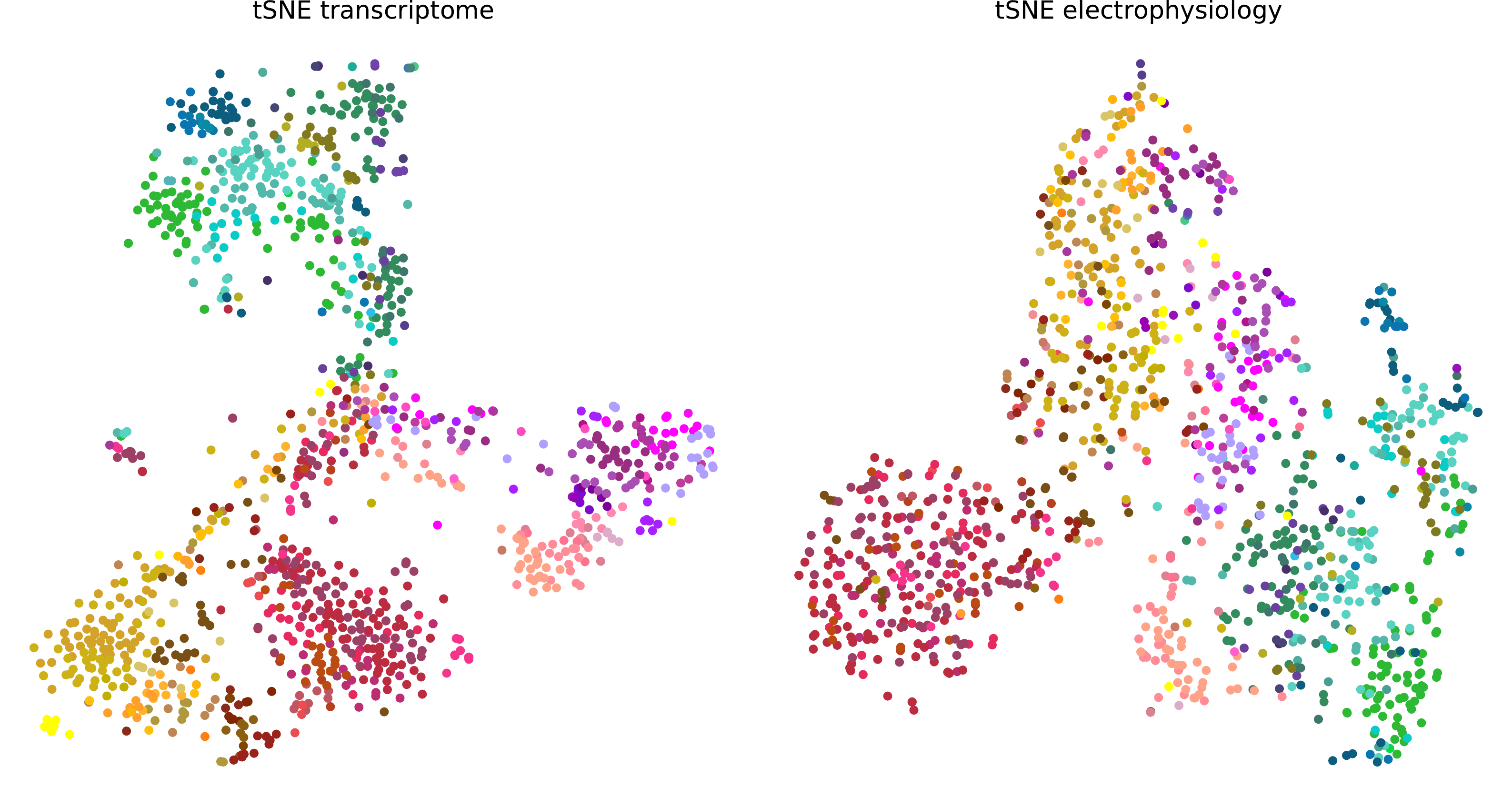}
\caption{Separate visualizations of two modalities in the M1 dataset. \textbf{Left} 
t-SNE of the transciptomic space. \textbf{Right} t-SNE of the electrophysiological space. The \texttt{openTSNE} package was used.}
\label{fig:scala_tsne}
\end{figure}

\begin{figure}[ht]
\centering
\includegraphics[width=\linewidth]{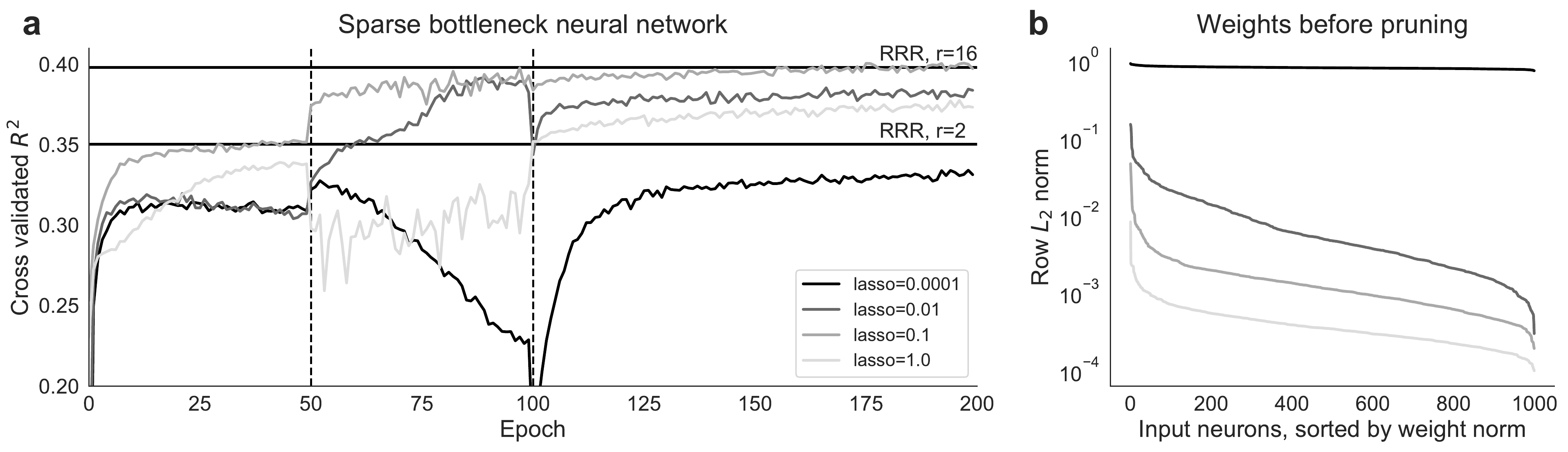}
\caption{The effect of the lasso penalty on the M1 dataset. \textbf{(a)} Validation (solid lines) $R^2$ of the sBNN-2 model, depending on the value of the lasso penalty (color-coded, see legend). The bottom two layers are frozen for 50 epochs after pre-training, then unfrozen for another 50 epochs. All models were pruned to 25 input units at epoch 100 and trained for further 100 epochs. Horizontal lines show maximum performances of rank-2 and full-rank sRRR with 25 genes. \textbf{(b)} The $\ell_2$ norms of the first layer weights for each input units just before pruning, depending on the value of the lasso penalty. These models were trained on the entire dataset. Vertical axis is on the log scale.}
\label{fig:different_lasso}
\end{figure}

\end{document}